%% file: main.tex
\documentclass[runningheads]{llncs}
\usepackage[T1]{fontenc}
\usepackage{times}
\usepackage{helvet}
\usepackage{courier}
\usepackage{xcolor}
\usepackage{amsmath}
\usepackage{natbib} 
\usepackage{amssymb}
\usepackage{multirow}
\usepackage{url}
\usepackage{subcaption}  
\usepackage{graphicx}
\usepackage{colortbl}
\usepackage{graphicx} 
\usepackage{booktabs} 
\usepackage[dvipsnames]{xcolor}
\usepackage[most]{tcolorbox}
\usepackage{xcolor}
\usepackage{graphicx}
\usepackage{lipsum} 

\begin{document}
%
\title{KGQuest: Template-Driven QA Generation from Knowledge Graphs with LLM-Based Refinement}
\titlerunning{KGQuest: Template-Driven QA Generation from KG and LLM}

\author{Sania Nayab\inst{1}\orcidID{0000-0002-3898-9091} \and
Marco Simoni\inst{2,3}\orcidID{0009-0000-4170-503X} \and
Giulio Rossolini\inst{1}\orcidID{0000-0002-6404-2627} \and
Andrea Saracino\inst{1,3}\orcidID{0000-0001-8149-9322}}

\newcommand {\del}[1]{{\color{red}\sf{[GR: #1]}}}

\authorrunning{S. Nayab et al.} 

\institute{
Scuola Superiore Sant’Anna, Pisa, Italy \and 
Sapienza University of Rome, Rome, Italy \and 
Institute of Informatics and Telematics, National Research Council of Italy (CNR)}
%
\maketitle              
\begin{abstract}
The generation of questions and answers (QA) from knowledge graphs (KG) plays a crucial role in the development and testing of educational platforms, dissemination tools, and large language models (LLM). However, existing approaches often struggle with scalability, linguistic quality, and factual consistency. This paper presents a scalable and deterministic pipeline for generating natural language QA from KGs, with an additional refinement step using LLMs to further enhance linguistic quality. The approach first clusters KG triplets based on their relations, creating reusable templates through natural language rules derived from the entity types of objects and relations. A module then leverages LLMs to refine these templates, improving clarity and coherence while preserving factual accuracy. Finally, the instantiation of answer options is achieved through a selection strategy that introduces distractors from the KG.
Our experiments demonstrate that this hybrid approach efficiently generates high-quality QA pairs, combining scalability with fluency and linguistic precision.

\end{abstract}
%
%
%



\input{intro_new}

\input{relwork}

\input{methodology_new2}
\input{experiments_new}
\input{conclusion}

\newpage
\input{appendix}

\bigskip
\bibliographystyle{plainnat}
\bibliography{main}
\end{document}

%% file: intro_new.tex
\section{Introduction}

Question–answer (QA) datasets and benchmarks have become foundational requirements for educational platforms, public knowledge dissemination, and training and evaluation of large language models (LLM) \cite{athreya2021template, formica2024template}. As much of today’s information is embedded in the form of knowledge graphs (KGs), whether general-purpose or domain-specific, effective strategies are needed to extract QA pairs that can be use for the above scopes. 
To reduce manual effort and address the scalability challenges inherent in this process, a wide range of methods have emerged over the past decade to automate QA generation from KGs \cite{Wang2022KGQG,saxena2021question,Ko2024NLFramework,zheng2015build}, ranging from early query-based and template-driven techniques to recent LLM-powered generative approaches.

While traditional methods are interpretable, they often rely on KG-specific query languages (e.g., SPARQL) \cite{abujabal2017automated}, which limits their portability and scalability across datasets and domains. Moreover, the rigidity of their templates, although providing deterministic outcomes, can sometimes produce ungrammatical or semantically awkward outputs, reducing overall linguistic quality \cite{athreya2021template, banerjee2023dblp}. In contrast, recent LLM-based approaches offer high fluency and adaptability in natural language generation but may struggle with interpretability when tasked with sentence formulation. In fact, as is well known, LLMs suffer from issues such as hallucination, bias amplification, and the risk of content collapse through recursive auto-generation cycles, which can also propagate into the generated datasets \cite{hu2023empirical, wu2023retrieve}.


\begin{figure*}[t]
\centering
\includegraphics[width=\textwidth]{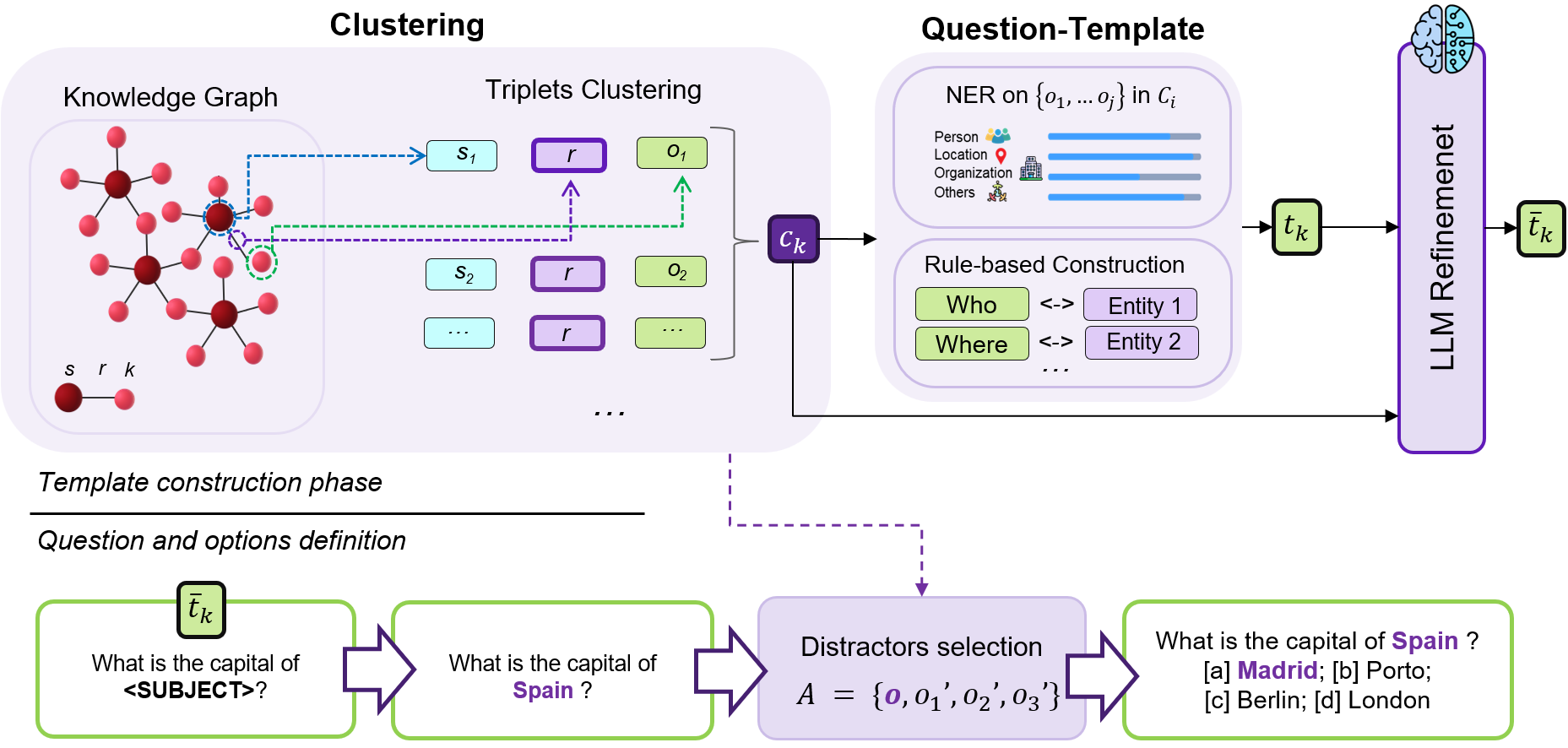}
\caption{\small{Overview of the QA generation pipeline. The top part illustrates the steps applied to extract a template (\(\bar{t}_k\)) from triplets in the knowledge graph. This includes a clustering process, the use of deterministic sentence-construction rules, and an LLM-based refinement step. The bottom part shows the instantiation of a triplet-based question \(\bar{q}\) from a refined template \(\bar{t}_k\), along with the computation of distractors to define the full set of answer options.}}
\label{f:pipeline}
\end{figure*}

Taking inspiration from the strengths and limitations of prior methods, we propose a unified and scalable QA generation paradigm. Our approach combines the controllability and determinism of natural-language templates with the support of LLMs only in a final refinement stage.
The pipeline works as follows: knowledge graph triplets (subject, predicate, object) are grouped into clusters sharing the same relation. Each cluster is then associated with a natural-language \emph{template} (e.g., \texttt{What is the capital of <SUBJECT>?}). Using a template, a preliminary QA instance is created: the object provides the correct answer, while plausible \textit{distractors} are mined from the KG through a selection strategy. Since all decisions are graph-driven and template-based, this stage remains reproducible and free from generative AI involvement.

While deterministic template generation ensures efficiency and factual consistency, it may still produce minor grammatical or stylistic inconsistencies, mainly due to the difficulty of covering diverse linguistic structures through static natural language rules.
To address this, we introduce an \emph{LLM-based template refinement} step, in which a reference sample from each template is analyzed by an LLM to identify and correct potential structural issues. The refined output is then generalized back into a template form, enabling systematic adjustments to the question structure. Importantly, this refinement requires only a single LLM inference per template, thereby preserving the factual integrity (i.e., avoiding the risk of introducing unpredictable triplet-specific content) and maintaining scalability.
An illustration of the approach is shown in Figure \ref{f:pipeline}.


We conduct evaluations to assess the quality of the generated QA pairs, with and without the refinement step. We also analyze the efficiency of applying LLMs at the template level rather than over the entire QA benchmark. Results confirm that our strategy yields a scalable, interpretable, and efficient use of LLMs for QA generation framework, particularly suited to contexts where transparency, fairness, and reliability are critical. Code of the framework is provided\footnote{The project code will be made available in a public repository upon acceptance.}.
%


To summarize, the main contributions of this work consist in the definition of a framework that introduces a first formulation of question templates extracted from knowledge graphs through a deterministic, rule-based pipeline for generating multiple-choice QA benchmarks. In addition, we propose a lightweight refinement step that leverages LLMs to improve these templates, enhancing linguistic quality at minimal cost. Furthermore, we provide a comprehensive evaluation of question quality, the refinement phase, and the efficiency of template-level LLM usage.





%% file: relwork.tex
\section{Related Work}
\label{relwork}
\paragraph{\textit{KG and QA:}} Knowledge graphs have long served as a robust foundation for question answering, particularly for fact-based queries. Early research primarily focused on parsing natural language questions into formal queries (e.g., SPARQL) over KGs such as Freebase \cite{Bollacker2008FreebaseAC}, QUINT \cite{abujabal2017automated}, and DBpedia \cite{lehmann2015dbpedia}. These systems emphasized accurate and complete retrieval of facts but were not designed to generate novel QA examples.


To overcome these challenges, neural network-based models have emerged as promising alternatives for generating diverse and natural questions directly from KGs. Models like RNN-based question generation \cite{indurthi2017generating}, and Graph2Seq \cite{wang2022generating} use attention mechanisms and graph encodings \cite{kacupaj2021conversational} to generate fluent questions. Moreover, approaches based on transformer models were used to generate QAs from KGs using their deep embeddings \cite{koncel2019text, han2022generating, saxena2021cronquestions}. 

While enhancing expressiveness and linguistic diversity, these approaches require large annotated datasets, consume significant computational resources, and are prone to generating hallucinated questions. This trade-off raises important questions about whether we truly need such heavy models or whether we can achieve scalable and trustworthy QA generation from the KG.

\paragraph{\textit{LLM based QA Generation}:} With the rise of LLMs, there is growing interest in generating QA datasets directly from text or structured sources using prompting-based techniques \cite{zhou2019multi, ko2024natural, hu2023empirical}. These methods have significantly impacted the field by enabling the generation of fluent, contextually rich questions and answers through prompt engineering \cite{zhang2023structure, liang2021knowledge}, with \citet{jiang2023structure} integrating scholarly KGs into LLM prompts to generate executable SPARQL queries.

Despite their generative strengths, LLMs are increasingly used as assistive components in hybrid pipelines, enhancing QA generation with knowledge graphs for tasks like grammar correction, paraphrasing, or template refinement \cite{pan2024unifying, guo2024improving, liu2024zvqaf}.

Very few studies \cite{rodriguez2022end, chomphooyod2023english} unify LLM capabilities with KG question answers. This hybrid strategy leverages the strengths of both deterministic KG-based generation and LLM-based linguistic enhancement, such as multi-hop KG reasoning \cite{yasunaga2021qa}, feedback-based QA generation \cite{kaiser2021reinforcement}, and linguistic refinement for answer over KG \cite{chakraborty2024multi}. While these hybrid systems help improve fluency and accuracy, they still rely on computationally expensive LLMs, and may lack the scalability and efficiency needed for large-scale, fact-aligned QA generation.

\paragraph{\textbf{This work}} Our approach bridges the gap between scalable, fact-aligned generation and natural language quality. Unlike fully generative LLM-based pipelines or rigid template-only methods, we introduce a lightweight, reproducible and scalable pipeline for generating QA benchmarks directly from KGs, with an optional refinement module that employs open-source LLMs only at the predicate category level. This approach includes an optional refinement module that uses small LLMs for question generation at the template \(t_k\) question level, ensuring semantic accuracy, including grammar, syntax, and formatting errors, linguistic clarity, and efficiency without compromising transparency or computational cost.

%% file: methodology_new2.tex
\section{Methodology}
\label{methodology}
This section outlines the methodology starting from background and overview for generating multiple-choice QA pairs from a knowledge graph. The approach mostly relies on small LLMs for question synthesis and is optimized for scalability and transparency, relying on the use of templates.

\subsection{Background and Overview}
We consider a knowledge graph \( G \), consisting of factual triplets \( (s, r, o) \), where \( s \in E \) is a subject, \( r \in R \) is a predicate, and \( o \in E \) is an object. $E$ and $R$ represent respectively the set of entities and predicate in \( G \).  The goal is to generate multiple-choice QA pairs, with the question derived from \( s \) and \( r \), the correct answer from \( o \), and other options (i.e., distractors) selected from entities \( o' \in E \setminus \{o\} \) such that \( (s, r, o') \notin G \).

Conceptually, the proposed pipeline, illustrated in Figure~\ref{f:pipeline}, first organizes all triplets \( (s, r, o) \in \mathcal{G} \) into clusters by grouping those that share the same relation \( r \). This process yields a set of clusters \( \mathcal{C} = \{ c_1, \dots, c_k, \dots, c_{|\mathcal{C}|} \} \). For each cluster \( c_k \in \mathcal{C} \), a predefined natural language question template \( t_k \) is assigned, encoding the structure of the question to be generated according to the shared relation \( r \) and the entities populating \( c_k \). 
For example, a possible template is \texttt{"What is the capital of <SUBJECT>?"}, obtained from a cluster \( c_k \) where \( r = \texttt{CapitalOf} \). The corresponding object (e.g., \texttt{Madrid}) represents the correct answer to be included among the options, while also serving to extract the predicate of the template. The obtained templates are then refined by the LLM, denoted as \( \bar{t}_k \), to correct potential grammatical errors and revise their formulation (this process is discussed in Section~\ref{refineLLM}). 
Finally, given a specific triplet in \( c_k \), the question $\bar{q}$ is instantiated by filling the placeholders in \( \bar{t}_k \) with the subject \( s \) from the triplet \( (s, r, o) \). Concerning the answer options appended to the generated question, the correct answer is provided by the object \( o \), while the distractor set is constructed by selecting entities \( o' \in E \setminus \{o\} \) (see Section \ref{ss:distractors}).


\subsection{Triplet Clustering and Template Construction}
\label{sec:template_construction}
To ensure consistent and scalable generation of natural language questions, factual triplets \( (s, r, o) \in \mathcal{G} \) are organized into clusters \(\mathcal{C}\), grouped according to relation \(r\), which is crucial for understanding the high-level semantic structure the question template of cluster \( c_k \) must have.

To first define a unified template form that can be applied across all triplets in cluster \(c_k\), also a reference object type $\mathcal{E}$ is extracted. Specifically, the object type influences the phrasing style of the question. For instance, if it is a \texttt{'PersonName'}, the sentence will most likely start with \texttt{'Who'}. Thus, a \text{named entity recognition} analysis is applied to the triplets in the cluster, and based on this distribution, the most common object type $\mathcal{E}_k$ is selected for constructing the template \(t_k\). Then, predefined and deterministic rules are then used to construct the template based on the relation \(r\) and $\mathcal{E}_k$. These rules consist of a simple mapping between entity types and question prefixes, as in the example above, together with a verbalized form of the relation \(r\). Please note that the list of mapping rules is fully deterministic and customizable, within the proposed framework.
Finally a subject placeholder \texttt{<SUBJECT>} is placed at the end of the question.

Note that the rationale of using simple and efficient mapping rules enables the generation of quick templates, which can be refined later by the LLM-refinement without introducing factual inconsistencies.



\subsection{Template Refinement Based on LLMs}
\label{refineLLM}

\begin{figure}[t]
\centering
\includegraphics[width=\columnwidth]{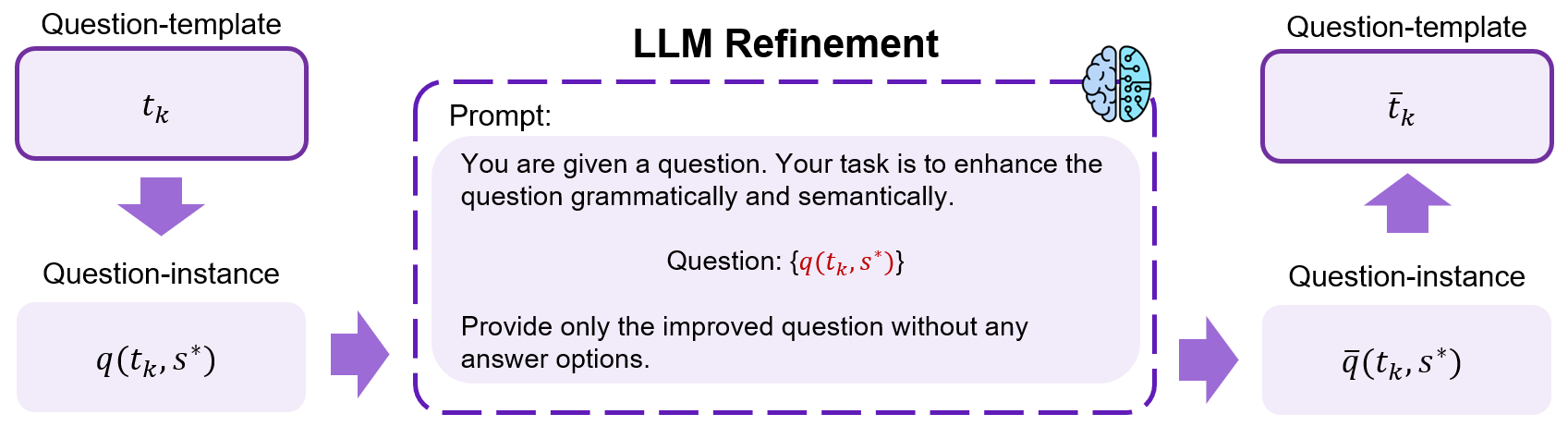}
\caption{\small{Overview of the LLM-based Template Refinement Process. An LLM is applied to refine a given template \(t_k\), correcting potential grammatical errors }} 
\label{f:reflection}
\end{figure}

Given a question template $t_k$ derived from the simple yet scalable approach discussed above, a refinement step is applied at the template level (to preserve efficiency) using an off-the-shelf LLM. The rationale for adopting LLMs at this stage is to leverage their extensive natural language knowledge to refine questions, while avoiding the unpredictability of generating them entirely from scratch.  

As shown in Figure~\ref{f:reflection}, the LLM-based template refinement proceeds as follows. First, a representative sample from the template cluster \( c_k \) is selected. In particular, a random sampling process extracts a triplet \((s^*, r, o^*)\), where the type of \( o^* \) is constrained to match the most common entity type \( \mathcal{E}_k \). The subject of this triplet, \( s^* \), is then used to instantiate a question from the template:  
\( q(t_k, s^*). \)
Next, this instantiated question is provided as input to an off-the-shelf LLM, which returns a refined version  
\( \bar{q}(t_k, s^*). \)
The specific prompting strategy used for the refinement (See Figure~\ref{f:reflection}), where the LLM is instructed to produce directly an improved version of the question while omitting other details. The resulting refined question is denoted as \( \bar{q} \).
Finally, \( s^* \) in \(\bar{q}(t_k, s^*)\) is identified and replaced with the tag \texttt{<subject>}, thereby restoring the template form and yielding a refined version \( \bar{t}_k.\)

\subsection{Question Instantiation and Distractors Selection}
\label{ss:distractors}
Given a triplet \( (s, r, o) \in c_k \) and its corresponding refined template \( \bar{t}_k \), a natural language question \(\bar{q}(\bar{t}_k, s)\) is instantiated online by replacing the subject tag in \( \bar{t}_k \) with the subject \(s\). This process is illustrated in the last part of Figure~\ref{f:pipeline}.

Once the question $\bar{q}$ is generated, the final step is the selection of distractors, which, as a good practice, should be incorrect yet plausible and contextually relevant alternatives. 
In the proposed framework, distractors are defined as any entities \( o' \in E \setminus \{o\} \) such that the triplet \((s, r, o') \notin G\). This ensures that each object \(o'\) represents a wrong option, assuming the knowledge graph is an oracle set of information.  
Note that, beyond the proposed strategy, ad hoc selection methods can be used to adjust the difficulty and complexity of the distractors \cite{nayab2025leveraging}.

In our analysis, we randomly selected \(N-1\) distractor objects,  
$D = \{o'_1, \dots, o'_{N-1}\},$
from the same cluster \( c_k \) of the original triplet, i.e., sharing the same relation \(r\). 
This simple yet effective process ensures that the distractors are semantically relevant, linguistically compatible, and contextually appropriate. 
Based on this selection process, the final set of \(N\) textual answer choices \(\mathcal{A}\) is formed by combining the correct answer \(o\) with the selected distractors, i.e., $\mathcal{A} = \{ o \} \cup D$.

%% file: experiments_new.tex
\section{Experiments}
\label{experiments}
In this section, after describing the experimental setup, we present results demonstrating the effectiveness of the proposed pipeline, focusing on its efficiency and reliability in preventing potential errors both before and after the LLM-based refinement step.

\paragraph{\textbf{Experimental setup.}}
To demonstrate the generalizability of our approach across different domains, we rely on three publicly available and widely used KGs: Wikigraphs \cite{wang2021wikigraphs}, WebQSP \cite{luo2023reasoning}, and CWQ \cite{luo2023reasoning}. These benchmarks have been selected as they cover different domains and scales, allowing us to test the robustness of our pipeline both on large, heterogeneous graphs and on more compact, domain-specific resources. For each of these KGs, we generate templates following the proposed pipeline and evaluate the resulting questions. Regarding the template addressed for generating questions, both $t_k$ and $\bar{t}_k$ (refined version) are assessed in the experiments to independently measure the quality of each step (in Section \ref{twofoldjudge} and \ref{sec:refinement}, respectively).

The three knowledge graphs (KGs) used in our experiments differ considerably in size with respect to the number of triplets: Wikigraphs includes approximately 367K triplets, WebQSP comprises around 18K triplets, and CWQ contains about 37K triplets. This variation ensures that we can evaluate how the method scales as the graph size grows. It is important to note that the maximum number of questions that can be generated from each graph matches the number of triplets, which highlights the potential of our approach to produce hundreds of thousands of questions in large-scale scenarios.

For implementing the LLM-based refinement stage (Section~\ref{refineLLM}), we compare the use of several efficient yet effective LLMs, specifically LLaMA3.2-70B, LLaMA3-8B, LLaMA3-3B \cite{touvron2023llama}, Phi-3.5 \cite{abdin2024phi}, Gemma2-2B \cite{team2024gemma}, and Qwen2.5-0.5B \cite{bai2023qwen}. The inclusion of models with different sizes and training philosophies allows us to investigate whether the refinement task truly requires very large LLMs or can be reliably addressed by smaller and more lightweight models. Following the proposed procedure, the LLM is employed to refine a template; subsequently, the refined template $\bar{t_k}$ is evaluated on a different sample from the same cluster (distinct from the one used for refining the original template $t_k$). This separation ensures that the evaluation does not trivially benefit from overlap with the refinement input, providing a more rigorous measure of generalization. 

For the evaluation of correctness and potential error types of the generated questions (in both Section \ref{twofoldjudge} and Section \ref{sec:refinement}), we adopt the \textit{LLMs-as-Judge} paradigm \cite{li2024llms, verga2024replacing}. In this setting, the quality of a question is determined by a \textit{jury} of three LLMs, applying the majority vote principle to mitigate the bias or inconsistency of any individual model. The jury evaluates one randomly selected question per cluster, which makes the evaluation computationally feasible while still covering a broad spectrum of generated questions. Specifically, we employ three state-of-the-art LLMs as judges: LLaMA3.3-70B \cite{touvron2023llama}, Qwen2-72B \cite{bai2023qwen}, and Phi-4 \cite{abdin2024phi}. We then report the proportion of questions judged as correct, as well as those flagged for grammatical, formatting, or syntactic errors.

It is important to note that the LLMs used as jury members differ from those adopted in the refinement stage, including also LLaMA-70B, which belongs to the same model family but uses a different release version. The rationale is twofold: first, to ensure fairness in the evaluation process and avoid circularity; and second, to highlight that the jury relies on significantly larger and more accurate LLMs, as indicated by their size and training configuration. This design demonstrates that, as confirmed by our results (Section \ref{sec:refinement}), the adoption of smaller LLMs in the refinement stage is sufficient to address the task effectively, without requiring computationally intensive large models.

Finally, all the prompts adopted for the use of LLMs, both for the jury evaluation and for directly generating questions from triplets (used for comparisons, e.g., Figure \ref{fig:examples_cwq}), are reported in the Appendix, to ensure full reproducibility of our experiments.

\subsection{Evaluation of the Templated-based Step}
\label{twofoldjudge}

\begin{figure*}[t]
\centering
\includegraphics[width=\textwidth]{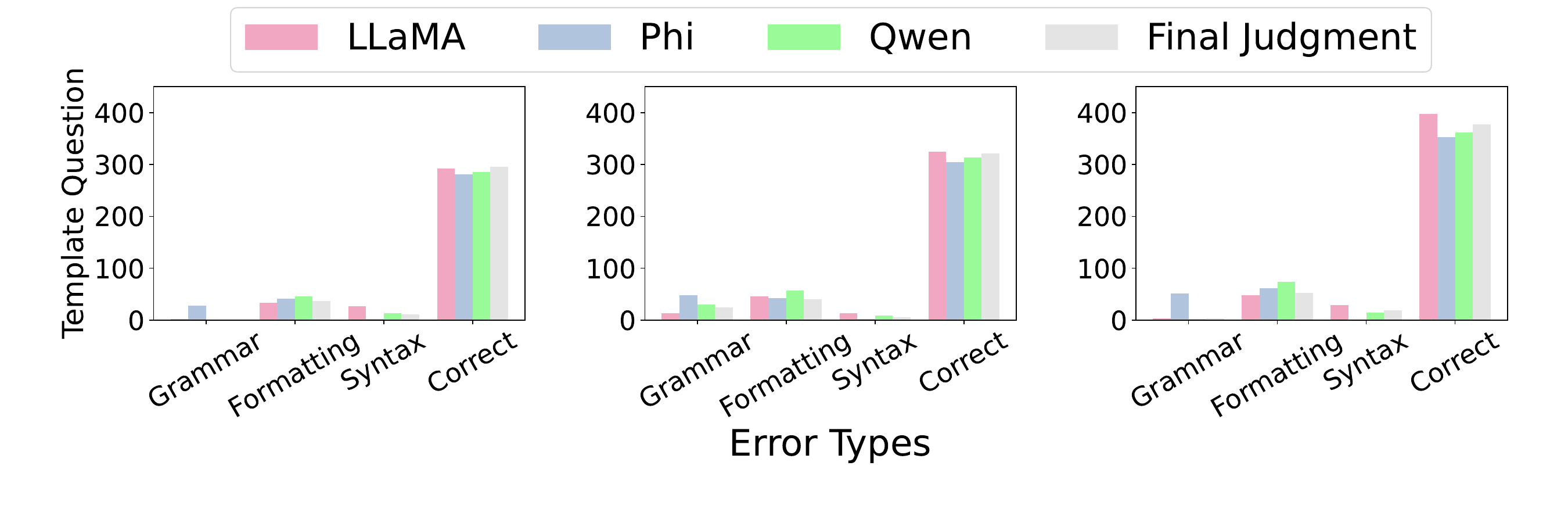}
\vspace{-2.5em}
\caption{\small{Results by Error Type for Question instantiated from the template \(t_k\) for the Wikigraphs, WebQSP, and CWQ KG datasets, respectively are shown in the subplots. The figures illustrate the distribution of error types (Grammar, Formatting, Syntax, Correct) for the LLaMA, Phi, and Qwen models, along with the final judgments.}}
\label{fig:error_results}
\end{figure*}

In this section, we evaluate the template pipeline construction defined in Section \ref{sec:template_construction}. Specifically, all the questions generated using the template pipeline were examined by the Jury, which assigned each one a label from the following four: correct, grammar error, formatting error, or syntax error. Figure~\ref{fig:error_results} shows the judgments of each individual judge, along with the final majority vote of the jury across the three knowledge graphs. Overall, the template-based generated questions performs well: between 80–90\% of the generated questions are considered correct, confirming that the template approach produces mostly well-formed instances. The majority of errors concern formatting issues such as punctuation, capitalization, or spacing, while grammatical and syntactic problems are far less frequent. Considering the jury judgement (in grey), syntax errors appear only in CWQ and Wikigraphs, and grammar errors are detected mainly in WebQSP. This suggests that the template-based pipeline is able to capture the linguistic structure of questions reliably, with most of the remaining issues limited to minor inconsistencies. Looking at the KGs in more detail, for CWQ the pipeline achieves the highest level of correctness, with very few grammatical or syntactic issues but still some formatting inconsistencies. For WebQSP the pipeline shows more grammar-related errors, likely due to the presence of relations that require subtler morphosyntactic agreements. For Wikigraphs, instead, the pipeline exhibits more syntax errors.


\subsection{Evaluation of the Refined Questions with LLMs}
\label{sec:refinement}

\input{tables/Ablation_Error}

In this section, we evaluate the questions obtained by refining, through LLMs, the templates generated in the template construction pipeline.  The results of this evaluation, shown in Table~\ref{t:LLMAugmentation_ablation}, confirm the effectiveness of this refinement step. Across all three KGs, grammar, formatting are removed, and syntax errors also decrease significantly. The proportion of correct questions judged by the jury increases accordingly: for example, CWQ reaches up to 373 correct questions, WebQSP up to 388, and Wikigraphs up to 345, all improvements compared to the pipeline-only evaluation.

One of the most important aspects highlighted by Table~\ref{t:LLMAugmentation_ablation} is that we improved the templates using LLMs independently of their size. In fact, smaller LLMs seem to make fewer mistakes than LLaMA3.2 70B. This means that in order to improve the quality of the templates and consequently the quality of questions, small LLMs can be sufficient, improving the efficiency of the entire process and reducing the required resources. 

\paragraph{Qualitative Analysis of generated questions.}

\begin{figure*}[!ht]
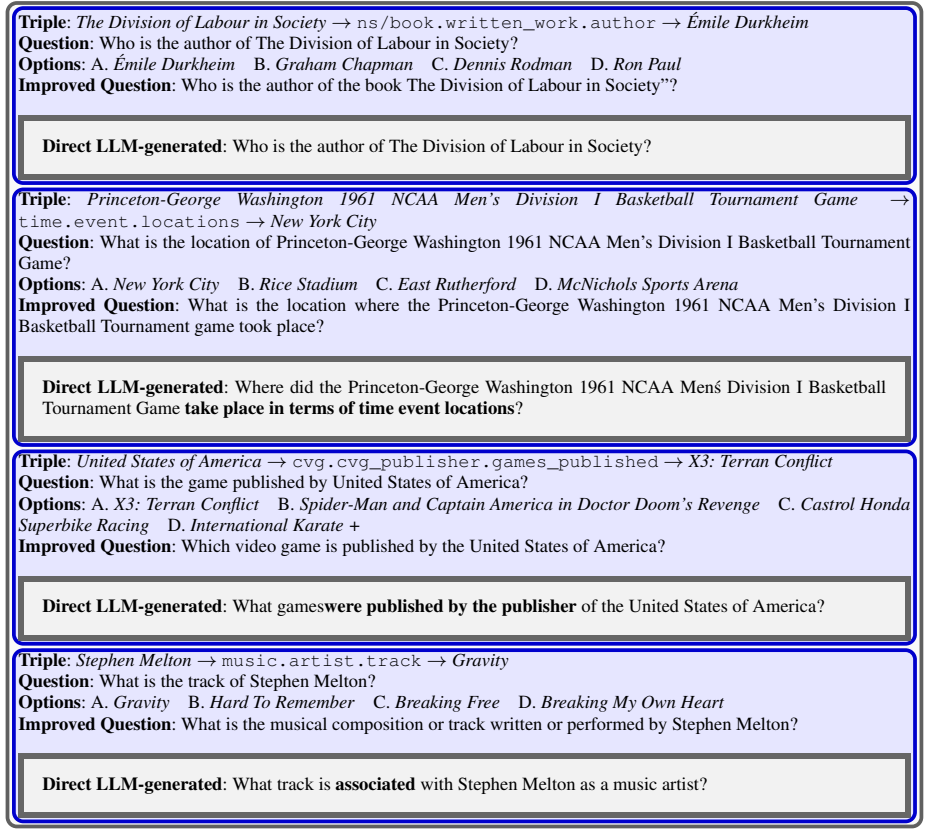

\begin{minipage}{\textwidth}
\begin{tcolorbox}[
    colback=gray!5!white,
    colframe=gray!75!black,
    fonttitle=\scriptsize,
    boxsep=0pt, left=1pt, right=1pt, top=0.5pt, bottom=0.5pt,
    coltitle=black,
    enhanced,
    before skip=0pt, after skip=0pt
]
\scriptsize

\begin{tcolorbox}[
    colback=blue!10!white,
    colframe=blue!80!black,
    fonttitle=\scriptsize,
    boxsep=0pt, left=1pt, right=1pt, top=0.5pt, bottom=0.5pt,
    coltitle=black,
    enhanced,
    before skip=1pt, after skip=1pt
]
\scriptsize
\textbf{Triple}: \textit{The Division of Labour in Society} $\rightarrow$ \texttt{ns/book.written\_work.author} $\rightarrow$ \textit{Émile Durkheim}\\
\textbf{Question}: Who is the author of The Division of Labour in Society?\\
\textbf{Options}: A. \textit{Émile Durkheim}\quad B. \textit{Graham Chapman}\quad C. \textit{Dennis Rodman}\quad D. \textit{Ron Paul}\\
\textbf{Improved Question}: Who is the author of the book The Division of Labour in Society''? \\
\begin{tcolorbox}[colback=gray!10!white, colframe=gray!80!black, boxrule=0.8mm, sharp corners=all, coltitle=black, fonttitle=\small, left=4pt, right=4pt, top=4pt, bottom=4pt, before skip=1pt, after skip=1pt]
\textbf{Direct LLM-generated}: Who is the author of The Division of Labour in Society?
\end{tcolorbox}
\end{tcolorbox}

\begin{tcolorbox}[
    colback=blue!10!white,
    colframe=blue!80!black,
    fonttitle=\scriptsize,
    boxsep=0pt, left=1pt, right=1pt, top=0.5pt, bottom=0.5pt,
    coltitle=black,
    enhanced,
    before skip=1pt, after skip=1pt
]
\scriptsize
\textbf{Triple}: \textit{Princeton-George Washington 1961 NCAA Men's Division I Basketball Tournament Game } $\rightarrow$ \texttt{time.event.locations} $\rightarrow$ \textit{New York City}\\
\textbf{Question}: What is the location of Princeton-George Washington 1961 NCAA Men's Division I Basketball Tournament Game?\\
\textbf{Options}: A. \textit{New York City}\quad B. \textit{Rice Stadium}\quad C. \textit{East Rutherford}\quad D. \textit{McNichols Sports Arena}\\
\textbf{Improved Question}: What is the location where the Princeton-George Washington 1961 NCAA Men's Division I Basketball Tournament game took place? \\
\begin{tcolorbox}[colback=gray!10!white, colframe=gray!80!black, boxrule=0.8mm, sharp corners=all, coltitle=black, fonttitle=\small, left=4pt, right=4pt, top=4pt, bottom=4pt, before skip=1pt, after skip=1pt]
\textbf{Direct LLM-generated}: Where did the Princeton-George Washington 1961 NCAA Men\'s Division I Basketball Tournament Game \textbf{take place in terms of time event locations}?
\end{tcolorbox}
\end{tcolorbox}

\begin{tcolorbox}[
    colback=blue!10!white,
    colframe=blue!80!black,
    fonttitle=\scriptsize,
    boxsep=0pt, left=1pt, right=1pt, top=0.5pt, bottom=0.5pt,
    coltitle=black,
    enhanced,
    before skip=1pt, after skip=1pt
]
\scriptsize
\textbf{Triple}: \textit{United States of America} $\rightarrow$ \texttt{cvg.cvg\_publisher.games\_published} $\rightarrow$ \textit{X3: Terran Conflict}\\
\textbf{Question}: What is the game published by United States of America?\\
\textbf{Options}: A. \textit{X3: Terran Conflict}\quad B. \textit{Spider-Man and Captain America in Doctor Doom's Revenge}\quad C. \textit{Castrol Honda Superbike Racing}\quad D. \textit{International Karate +}\\
\textbf{Improved Question}: Which video game is published by the United States of America? \\
\begin{tcolorbox}[colback=gray!10!white, colframe=gray!80!black, boxrule=0.8mm, sharp corners=all, coltitle=black, fonttitle=\small, left=4pt, right=4pt, top=4pt, bottom=4pt, before skip=1pt, after skip=1pt]
\textbf{Direct LLM-generated}: What games\textbf{were published by the publisher} of the United States of America?
\end{tcolorbox}
\end{tcolorbox}

\begin{tcolorbox}[
    colback=blue!10!white,
    colframe=blue!80!black,
    fonttitle=\scriptsize,
    boxsep=0pt, left=1pt, right=1pt, top=0.5pt, bottom=0.5pt,
    coltitle=black,
    enhanced,
    before skip=1pt, after skip=1pt
]
\scriptsize
\textbf{Triple}: \textit{Stephen Melton} $\rightarrow$ \texttt{music.artist.track} $\rightarrow$ \textit{Gravity}\\
\textbf{Question}: What is the track of Stephen Melton?\\
\textbf{Options}: A. \textit{Gravity}\quad B. \textit{Hard To Remember}\quad C. \textit{Breaking Free}\quad D. \textit{Breaking My Own Heart}\\
\textbf{Improved Question}: What is the musical composition or track written or performed by Stephen Melton? \\
\begin{tcolorbox}[colback=gray!10!white, colframe=gray!80!black, boxrule=0.8mm, sharp corners=all, coltitle=black, fonttitle=\small, left=4pt, right=4pt, top=4pt, bottom=4pt, before skip=1pt, after skip=1pt]
\textbf{Direct LLM-generated}: What track is \textbf{associated} with Stephen Melton as a music artist?
\end{tcolorbox}
\end{tcolorbox}

\end{tcolorbox}
\end{minipage}
\caption{\small{Examples of generated triplets with LLama-70b from the selected KG, along with the corresponding questions instantiated from templates, first extracted through the deterministic step and then refined. For comparison, questions generated directly by the LLM are also provided in grey, with potential hallucination issues highlighted in bold.}}
\label{fig:examples_cwq}
\end{figure*}

To better assess the quality of the generated questions, we provide illustrative examples in \ref{fig:examples_cwq}. In each example, we report the triplet used, the question produced by the deterministic template, the corresponding set of answer options, and the refined version obtained through the LLM-based template refinement step. For comparison, we also include, in grey, a question generated directly by prompting the same LLM (LLaMA-70B) to produce a question from the given triplet (see appendix for details).

This side-by-side comparison highlights two key aspects. First, the effect of the template refinement stage on the overall quality of the questions is relatively limited: the refinement does not alter the semantics of the question nor introduce additional knowledge, but rather improves the fluency, readability, and naturalness of the phrasing. In other words, the refined question remains fully faithful to the original template-based formulation, while appearing more aligned with human-authored language.

Second, the direct LLM-based generation from triplets often results in the inclusion of information that is not contained in the original data, thereby introducing hallucinations (highlighted in bold in the examples). 
This direct approach is computationally demanding, as the LLM is invoked at each question generation step, unlike our pipeline, where the LLM is applied only once per cluster during refinement. This makes our method considerably more efficient and scalable, while maintaining stronger guarantees of consistency with the source knowledge. 

\paragraph{Efficiency Analysis.}

 Table \ref{genTime} shows, depending on the KG, the time needed to refine the templates and the approximate time needed to generate questions directly from triples. Obviously, the latter is incredibly larger because the model needs to take all the triples that are present inside the KG; instead, our pipeline just needs to pass to the model one sample per cluster, so one sample per individual category.
 
\input{tables/InfTime}

The table shows that the difference in efficiency is not marginal but spans several orders of magnitude. For instance, on Wiki (367K questions), direct generation with LLaMA-70B would require more than \textbf{160 hours}, while our pipeline completes the refinement in less than 10 minutes. On smaller datasets, such as WebQSP, the overhead is still significant: direct generation demands up to 14 hours, compared to just a few minutes for refinement. Importantly, this efficiency holds across all tested models, from large-scale ones (LLaMA-70B) to lightweight alternatives (Qwen2.5-0.5B), confirming that the scalability of our approach does not depend on a specific architecture.

This efficiency translates directly into practical benefits: our pipeline makes large-scale question generation feasible in real scenarios, reducing both computational costs and wall-clock time, and enabling frequent updates of the generated datasets as the underlying KGs evolve.



%% file: tables/Ablation_Error.tex
\begin{table}[t]
\centering
\caption{\small{Jury evaluation of refined template question (Question instantiated from the template \(t_k\)) after the LLM-based refinements step. Distribution of error types (Grammar, Formatting, Syntax) and correct questions across datasets. }}
\label{t:LLMAugmentation_ablation}
\tiny
\resizebox{0.95\columnwidth}{!}{%
\begin{tabular}{l cccc cccc cccc}
\toprule
\multirow{2}{*}{\textbf{Model}} & 
\multicolumn{4}{c}{\textbf{Wiki}} & 
\multicolumn{4}{c}{\textbf{WebQSP}} & 
\multicolumn{4}{c}{\textbf{CWQ}} \\
\cmidrule(lr){2-5} \cmidrule(lr){6-9} \cmidrule(lr){10-13}
 & Gram. & Form. & Syntax & Correct 
 & Gram. & Form. & Syntax & Correct 
 & Gram. & Form. & Syntax & Correct \\
\midrule
LLaMA-70B & 0 \textcolor{ForestGreen}{/1} & 1 \textcolor{ForestGreen}{/36} & 1 \textcolor{ForestGreen}{/10} & 343 
          & 0 \textcolor{ForestGreen}{/24} & 2 \textcolor{ForestGreen}{/38} & 0 \textcolor{ForestGreen}{/6}  & 388 
          & 0 \textcolor{ForestGreen}{/3}  & 0 \textcolor{ForestGreen}{/53} & 1 \textcolor{ForestGreen}{/18} & 372 \\
LLaMA-8B  & 0 \textcolor{ForestGreen}{/1} & 1 \textcolor{ForestGreen}{/36} & 1 \textcolor{ForestGreen}{/10} & 343 
          & 0 \textcolor{ForestGreen}{/24} & 0 \textcolor{ForestGreen}{/40} & 1 \textcolor{ForestGreen}{/5}  & 387 
          & 0 \textcolor{ForestGreen}{/3}  & 0 \textcolor{ForestGreen}{/53} & 0 \textcolor{ForestGreen}{/19} & 373 \\
LLaMA-3B  & 0 \textcolor{ForestGreen}{/1} & 0 \textcolor{ForestGreen}{/37} & 0 \textcolor{ForestGreen}{/11} & 345 
          & 0 \textcolor{ForestGreen}{/24} & 1 \textcolor{ForestGreen}{/39} & 0 \textcolor{ForestGreen}{/6}  & 387 
          & 0 \textcolor{ForestGreen}{/3}  & 0 \textcolor{ForestGreen}{/53} & 0 \textcolor{ForestGreen}{/19} & 373 \\
Phi-3.5   & 0 \textcolor{ForestGreen}{/1} & 0 \textcolor{ForestGreen}{/37} & 0 \textcolor{ForestGreen}{/11} & 345 
          & 0 \textcolor{ForestGreen}{/24} & 0 \textcolor{ForestGreen}{/40} & 0 \textcolor{ForestGreen}{/6}  & 388 
          & 0 \textcolor{ForestGreen}{/3}  & 0 \textcolor{ForestGreen}{/53} & 0 \textcolor{ForestGreen}{/19} & 373 \\
Gemma-2   & 0 \textcolor{ForestGreen}{/1} & 0 \textcolor{ForestGreen}{/37} & 0 \textcolor{ForestGreen}{/11} & 345 
          & 0 \textcolor{ForestGreen}{/24} & 0 \textcolor{ForestGreen}{/40} & 1 \textcolor{ForestGreen}{/5}  & 387 
          & 0 \textcolor{ForestGreen}{/3}  & 0 \textcolor{ForestGreen}{/53} & 0 \textcolor{ForestGreen}{/19} & 373 \\
Qwen-0.5B & 0 \textcolor{ForestGreen}{/1} & 0 \textcolor{ForestGreen}{/37} & 0 \textcolor{ForestGreen}{/11} & 345 
          & 0 \textcolor{ForestGreen}{/24} & 0 \textcolor{ForestGreen}{/40} & 1 \textcolor{ForestGreen}{/5}  & 387 
          & 0 \textcolor{ForestGreen}{/3}  & 0 \textcolor{ForestGreen}{/53} & 0 \textcolor{ForestGreen}{/19} & 373 \\
\bottomrule
\end{tabular}
}
\end{table}

%% file: tables/InfTime.tex
\begin{table*}[t]
\centering
\scriptsize
\caption{\small{Comparison of time needed to refine the templates (blue) and to generate the questions from triples (red). Question number \textit{Qs} per KG (and so the number of triplets) is indicated between brackets. Time is represented in hours ('h') and mins ('m').}}
\label{genTime}
\tiny 
\resizebox{0.85\textwidth}{!}{%
\begin{tabular}{l|c|c|c}
\toprule
\textbf{Model} & \textbf{Wiki (345 cat. / 367K Qs)} & \textbf{CWQ (373 cat. / 18K Qs)} & \textbf{WebQSP (388 cat. / 37K Qs)} \\
\midrule

Llama3.3-70b 
& $\approx$ \textcolor{blue}{9 m} / \textcolor{red}{160 h} & $\approx$ \textcolor{blue}{9 m} / \textcolor{red}{7 h} & $\approx$ \textcolor{blue}{9 m} / \textcolor{red}{14 h} 
\\
\midrule

Llama3.1-8b 
& $\approx$ \textcolor{blue}{3 m} / \textcolor{red}{53 h} & $\approx$ \textcolor{blue}{3 m} / \textcolor{red}{3 h} & $\approx$ \textcolor{blue}{3 m} / \textcolor{red}{5 h} 
\\
\midrule

Llama3.2-3b 
& $\approx$ \textcolor{blue}{2 m} / \textcolor{red}{36 h} & $\approx$ \textcolor{blue}{3 m} / \textcolor{red}{3 h} & $\approx$ \textcolor{blue}{3 m} / \textcolor{red}{5 h} 
\\
\midrule

Phi-3.5-mini 
& $\approx$ \textcolor{blue}{6 m} / \textcolor{red}{106 h} & $\approx$ \textcolor{blue}{6 m} / \textcolor{red}{5 h} & $\approx$ \textcolor{blue}{6 m} / \textcolor{red}{10 h} 
\\
\midrule

Gemma2-2b 
& $\approx$ \textcolor{blue}{4 m} / \textcolor{red}{71 h} & $\approx$ \textcolor{blue}{5 m} / \textcolor{red}{4 h} & $\approx$ \textcolor{blue}{5 m} / \textcolor{red}{8 h} 
\\
\midrule

Qwen2.5-0.5B 
& $\approx$ \textcolor{blue}{2 m} / \textcolor{red}{36 h} & $\approx$ \textcolor{blue}{2 m} / \textcolor{red}{2 h} & $\approx$ \textcolor{blue}{2 m} / \textcolor{red}{3 h} 
\\
\bottomrule
\end{tabular}
}
\end{table*}

%% file: conclusion.tex
\section{Conclusion and Future Work}
\label{conclusion}


This work introduces a modular approach to automated QA generation from knowledge graphs, incorporating a step for improving linguistic clarity with LLMs. By structuring the pipeline around subject-relation-object clusters and reusable natural language templates, we aim to provide more transparent, reproducible, and scalable QA generation, addressing some of the limitations of traditional approaches. However, the framework can also generate questions that deviate from the template structure, potentially introducing semantic or formatting errors, as shown in the examples (see Figure \ref{fig:examples_cwq}).

The optional refinement module, based on off-the-shelf LLMs, mitigates formulation errors and enhances template-instanced question by improving fluency and correctness, while preserving factual integrity and minimizing computational costs. Empirical results (in Section \ref{sec:refinement}) confirm that this hybrid strategy offers measurable improvements in quality, with efficient inference performance. As LLMs are known to produce hallucinated generations, this can also be observed in the examples (see Figure \ref{fig:examples_cwq}), where LLMs associate attributes of the subject while refining the question template.

Future extensions of this work may explore several promising directions. One avenue is difficulty-aware QA generation, where distractor selection can be enhanced through knowledge-aware scoring or adaptive calibration strategies, enabling personalized or leveled question design. Additionally, exploring cross-domain QA generation could allow the framework to generalize to diverse knowledge graphs, expanding its applicability across different domains and tasks.

Overall, this approach rethinks QA generation as a controlled, template-first process, capable of producing high fidelity QA datasets suitable for educational, research, and benchmarking purposes across various domains.

%% file: appendix.tex
\section{Appendix}

\paragraph{Used prompt.}
In the following, we report the prompts employed for: (i) the evaluation of question quality by LLM judges; (ii) the LLM-based refinement stage; and (iii) the direct generation of questions from triplets (for comparative analysis in Figure~\ref{fig:examples_cwq} and Table~\ref{genTime}).

\begin{figure}[h]
\centering
\begin{minipage}{0.9\textwidth}
\begin{tcolorbox}[
    colback=gray!5!white,
    colframe=black!80!white,
    boxrule=0.6pt,
    arc=2pt,
    width=\textwidth,
    title={System Instruction and User Prompt for Judge Evaluation},
    fonttitle=\bfseries,
    sharp corners=south
]

\textbf{SYSTEM INSTRUCTION:}
You are an impartial judge responsible for evaluating the correctness of multiple-choice questions in terms of grammar, syntax, and formatting. Each question is generated from a structured triple consisting of three elements: \textit{subject}, \textit{relationship}, and \textit{object}. Your task is to ensure that the question correctly incorporates the subject and the relationship while \textbf{excluding} the object, as the object should only appear among the answer choices.

\textbf{Important Constraints:} (A) The \textbf{subject} must appear exactly as it is represented in the triple; (B) The \textbf{relationship} must be correctly integrated into the question; (C) The \textbf{object} must not appear in the question.
\vspace{0.1cm}
\textbf{Evaluation Criteria:}
\begin{enumerate}
    \item \textbf{Grammar}: Ensure proper grammatical rules.
    \item \textbf{Syntax}: Verify sentence structure.
    \item \textbf{Formatting}: Check answer choices for distinctness and correctness.
\end{enumerate}

\textbf{USER PROMPT:}
The following question has been generated from the triple of given \(t_k\): \texttt{\{category\}} 

\textbf{Question:} \texttt{\{question\}} 

Is this question correctly formulated?
\end{tcolorbox}
\end{minipage}
\end{figure}

\begin{figure}[h]
\centering
\begin{minipage}{0.9\textwidth}
\begin{tcolorbox}[
    colback=gray!5!white,
    colframe=black!80!white,
    boxrule=0.6pt,
    arc=2pt,
    width=\textwidth,
    title={System Instruction and User Prompt for Judge Evaluation},
    fonttitle=\bfseries,
    sharp corners=south
]

\noindent You are given one RDF-style triple formatted as:
\(\textit{subject} \;\rightarrow\; \texttt{relation} \;\rightarrow\; \textit{object}\)

\noindent\textbf{TRIPLE:} \texttt{\{triple\_str\}}

\noindent\textbf{TASK:} Write ONE natural-language question that: - Mentions the \textbf{SUBJECT} and encodes the \textbf{RELATION}; - Is answerable \textbf{ONLY} by the \textbf{OBJECT}; - Adds \textbf{NO extra facts, qualifiers, dates, or names} not present in the triple; - Keeps entities verbatim (do not rename, abbreviate, or translate them).

\noindent\textbf{FORMATTING:} - Output \textbf{ONLY the question text on one line ending with a question mark.}; - Do \textbf{NOT include the answer, labels, or any extra text};
\end{tcolorbox}
\end{minipage}
\end{figure}




